\documentclass{bmvc2k}

\usepackage{times}
\usepackage{epsfig}
\usepackage{graphicx}
\usepackage{amsmath}
\usepackage{amssymb}
\usepackage{multirow}
\usepackage{adjustbox}
\usepackage{diagbox}
\usepackage{threeparttable}

\def\ie{\emph{i.e.}}

\DeclareMathOperator{\upsample}{Upsample}
\newcommand{\img}[1]{\mathbf{#1}}
\newcommand{\R}{\mathbb{R}}

\newcommand{\wgd}[1]{\textcolor{black}{{}#1}}

\title{Student-Teacher Feature Pyramid Matching for Anomaly Detection}

\addauthor{Guodong Wang$^{\ast\mbox{\sffamily\scriptsize1,}}$}{wanggd@buaa.edu.cn}{2}
\addauthor{Shumin Han$^{\ast\mbox{\sffamily}}$}{hanshumin@baidu.com}{3}
\addauthor{Errui Ding}{dingerrui@baidu.com}{3}
\addauthor{Di Huang$^{\mbox{\sffamily\scriptsize \dag1,}}$}{dhuang@buaa.edu.cn}{2}

\addinstitution{
 State Key Laboratory of Software Development Environment\\
 Beihang University\\
 Beijing, China 
}
\addinstitution{
School of Computer Science and Engineering\\
Beihang University\\
Beijing, China
}
\addinstitution{
Department of Computer Vision Technology\\
Baidu, Inc.\\
Beijing, China
}

\runninghead{G. WANG ET AL}{STFPM ANOMALY DETECTION}

\begin{document}

\maketitle

\begin{abstract}
   Anomaly detection is a challenging task and usually formulated as an one-class learning problem for the unexpectedness of anomalies. This paper proposes a simple yet powerful approach to this issue, which is implemented in the student-teacher framework for its advantages but substantially extends it in terms of both accuracy and efficiency. Given a strong model pre-trained on image classification as the teacher, we distill the knowledge into a single student network with the identical architecture to learn the distribution of anomaly-free images and this one-step transfer preserves the crucial clues as much as possible. Moreover, we integrate the multi-scale feature matching strategy into the framework, and this hierarchical feature matching enables the student network to receive a mixture of multi-level knowledge from the feature pyramid under better supervision, thus allowing to detect anomalies of various sizes. The difference between feature pyramids generated by the two networks serves as a scoring function indicating the probability of anomaly occurring. Due to such operations, our approach achieves accurate and fast pixel-level anomaly detection. Very competitive results are delivered on the MVTec anomaly detection dataset, superior to the state of the art ones.
\end{abstract}

\section{Introduction}
\label{sec:intro}

Anomaly detection is generally referred to as identifying samples that are atypical with respect to regular patterns in the data set and has shown great potential in various real-world applications such as 
video surveillance~\cite{Abati2019,Roitberg2018}, 
product quality control~\cite{Bergmann2019,Bergmann2020,Napoletano2018} and 
medical diagnosis~\cite{Schlegl2019,Schlegl2017,Vasilev2018}. 
Its key challenge lies in the unexpectedness of anomalies which is very difficult to deal with in a supervised way, as labeling all types of anomalous instances seems unrealistic. 

\begin{figure}
	\centering
	\includegraphics[width=0.8\textwidth]{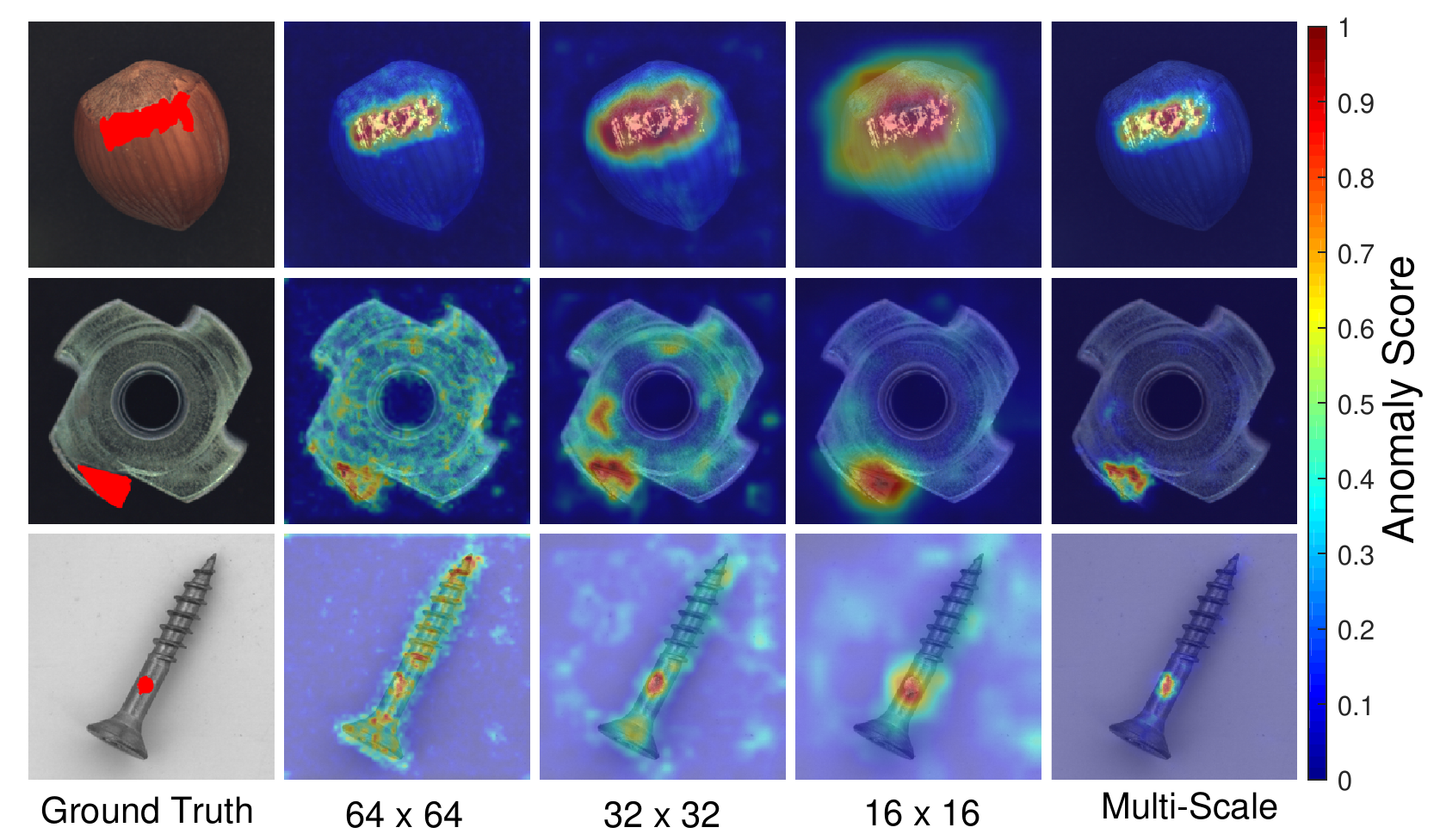}
	\caption{Visual results of our method on three defective images from the MVTec AD dataset. ResNet-18 is used as backbone and the three bottom blocks (\textit{i.e.}, conv2\_x, conv3\_x, conv4\_x) are selected as feature extractors. Columns from left to right correspond to input images with defects (ground truth regions in red), anomaly maps of the three blocks, and the resulting anomaly maps respectively.}
	\label{fig1}
\end{figure}

Previous studies address this challenge in the form of one-class learning paradigm~\cite{Moya1993}. They approximate the decision boundary for a binary classification problem by searching a feature space where the distribution of normal data is accurately modeled. Deep learning, in particular convolutional neural networks (CNNs)~\cite{Lecun1998} and residual networks (ResNets)~\cite{He2016}, provides a powerful alternative to automatically build comprehensive representations at multiple levels. Such deep features prove very effective in capturing the intrinsic characteristics of the normal data manifold~\cite{An2015,Chalapathy2018,Masana2018,Ruff2018,Zhou2017}. Despite the promising results in their respective fields, all these methods simply predict anomalies at the image-level without spatial localization.

The pixel-level methods advance anomaly detection by means of pixel-wise comparison of image patches and their reconstructions~\cite{Baur2018,Schlegl2019,Schlegl2017} or per-pixel estimation of probability density on entire images~\cite{Abati2019,Seebock2016}, among which Auto-encoders, Generative Adversarial Networks (GANs), and their variants are dominating models. However, their performance is prone to serious degradation when images are poorly reconstructed ~\cite{Bergmann2019b} or likelihoods are inaccurately calibrated~\cite{Nalisnick2019}.

Some recent attempts transfer the knowledge from other well-studied computer vision tasks. They directly apply the networks pre-trained on image classification and show that they are sufficiently generic to image-level detection~\cite{Andrews2016,Burlina2019,Erfani2016}. 
Cohen and Hoshen \cite{Cohen2020} investigate this idea in pixel-level detection and delivers performance gain; unfortunately, it has the time bottleneck due to per-pixel comparison. 
Bergmann~\textit{et al.}~\cite{Bergmann2020} utilize the pre-trained model in a more efficient way by implicitly learning the distribution of normal features with a student-teacher framework and reach decent results.
The difference between the outputs of the students and teacher along with the uncertainty among students' predictions serves as the anomaly scoring function. Nevertheless, two major drawbacks still remain: \ie,~the incompleteness of transferred knowledge and complexity of handling scaling. For the former, since knowledge is distilled from a ResNet-18~\cite{He2016} into a lightweight teacher network, the big gap between their model capacities~\cite{Wang2020} tends to incur loss of important information. For the latter, multiple student-teacher ensemble pairs are required to be separately trained, each for a specific respective field, to achieve scale invariance, which leads to the inconvenience in computation. Both the facts leave much room for improvement.

In this paper, we propose a simple yet powerful approach to anomaly detection, which follows the student-teacher framework for the advantages but substantially extends it in terms of both accuracy and efficiency. Specifically, given a powerful network pre-trained on image classification as the teacher, we distill the knowledge into a single student network with the identical architecture. In this case, the student network learns the distribution of anomaly-free images by matching their features with the counterparts of the pre-trained network, and this one-step transfer preserves the crucial information as much as possible. Furthermore, to enhance the scale robustness, we embed multi-scale feature matching into the network, and this hierarchical feature \wgd{matching} strategy enables the student network to receive a mixture of multi-level knowledge from the feature pyramid under a stronger supervision and thus allows to detect anomalies of various sizes (see~Figure \ref{fig1} for visualization). The feature pyramids from the teacher and student networks are compared for prediction, where a larger difference indicates a higher probability of anomaly occurrence.

Compared to the previous work, especially the preliminary student-teacher model, the benefits of our approach are two-fold. First, useful knowledge is well transferred from the pre-trained network to the student network within one-step distillation, as they share the same structure. Second, thanks to the hierarchical structure of the network, 
multi-scale anomaly detection is conveniently reached by the proposed feature pyramid matching scheme.
Due to such strengths, our approach conducts accurate and fast pixel-level anomaly detection. It reports very competitive results on the MVTec anomaly detection dataset, and more results on ShanghaiTech Campus (STC)~\cite{Luo_2017_ICCV} and CIFAR-10~\cite{Krizhevsky2009} are presented in the supplementary material.

\section{Related Work}
\label{sec2}

\subsection{Image-level Anomaly Detection}
\label{sec2-sub1}

Image-level techniques manifest anomalies in images of unseen categories. 
They can be coarsely divided into: reconstruction-based, distribution-based and classification-based.

The first group of approaches reconstruct the training images to capture the normal data manifold. An anomalous image is very likely to possess a high reconstruction error during inference, as it is drawn from a different distribution. The main weakness of these approaches comes from the excellent generalization ability of the deep models, including variational autoencoder~\cite{An2015}, robust autoencoder~\cite{Zhou2017}, conditional GAN~\cite{Akcay2018}, and bi-directional GAN~\cite{Zenati2018},  which probably allows anomalous images to be faithfully reconstructed. 

Distribution-based approaches model the probabilistic distribution of the normal images. The images that have low probability density values are designated as anomalous. Recent algorithms such as anomaly detection GAN (ADGAN)~\cite{Deecke2018} and deep autoencoding Gaussian mixture model (DAGMM)~\cite{Zong2018} learn a deep projection that maps high-dimensional images into a low-dimensional latent space. 
Nevertheless, these methods have high sample complexity and demand large training data.

Classification-based approaches have dominated anomaly detection in the last decade. One useful paradigm is to feed the deep features extracted by deep generative models~\cite{Burlina2019} or transferred from pre-trained networks~\cite{Andrews2016,Erfani2016} into a separate shallow classification model like one-class support vector machine (OC-SVM)~\cite{Scholkopf2001}. 
Another line of research depends on self-supervised learning. Geom~\cite{Golan2018} creates a dataset by applying dozens of geometric transformations to the normal images and trains a multi-class neural network over the self-labeled dataset to discriminate such transformations. At test time, anomalies are expected to be assigned with less confidence in discriminating the transformations. 

\subsection{Pixel-level Anomaly Detection}
\label{sec2-sub2}

Pixel-level techniques are particularly designed for anomaly localization. They aim to precisely segment anomalous regions in images, which is more complicated than binary classification.

The expressive power of deep neural networks inspires a series of studies that explore how to transfer the benefits of the networks pre-trained on image classification tasks to anomaly detection. Napoletano~\textit{et al.}~\cite{Napoletano2018} exploit a pre-trained ResNet-18 to embed cropped training image patches into a feature space, reduce the dimension of feature vectors by PCA, and model their distribution using K-means clustering. This method requires a large number of overlapping patches to obtain a spatial anomaly map at inference time, which results in coarse-grained maps and may become a performance bottleneck.

To avoid cropping image patches and accelerate feature extraction, Sabokrou~\textit{et al.}~\cite{Sabokrou2018} build descriptors from early feature maps of a pre-trained fully convolutional network (FCN) and adopt a unimodal Gaussian distribution to fit feature vectors of the anomaly-free images. 
\wgd{However, the unimodel Gaussian distribution fails to characterize the training feature distribution as the problem complexity increases.}
More recently, a convolutional adversarial variational autoencoder with guided attention (CAVGA)~\cite{shashanka2020attention} incorporates Grad-CAM~\cite{Selvaraju2017} into a variational autoencoder with an attention expansion loss to encourage the deep model itself to focus on all normal regions in the image. \wgd{Simliar to typical autoencoders (AE) \cite{Bergmann2019, Bergmann2019b} and variational autoencoders (VAE) \cite{liu2020towards}, CAVGA also suffers from the strong generalization ability which allows good reconstruction for anomalous images.}


\section{Method}
\label{sec3}


\subsection{Framework}

\begin{figure*}[!t]
	\centering
	\includegraphics[width=0.7\textwidth]{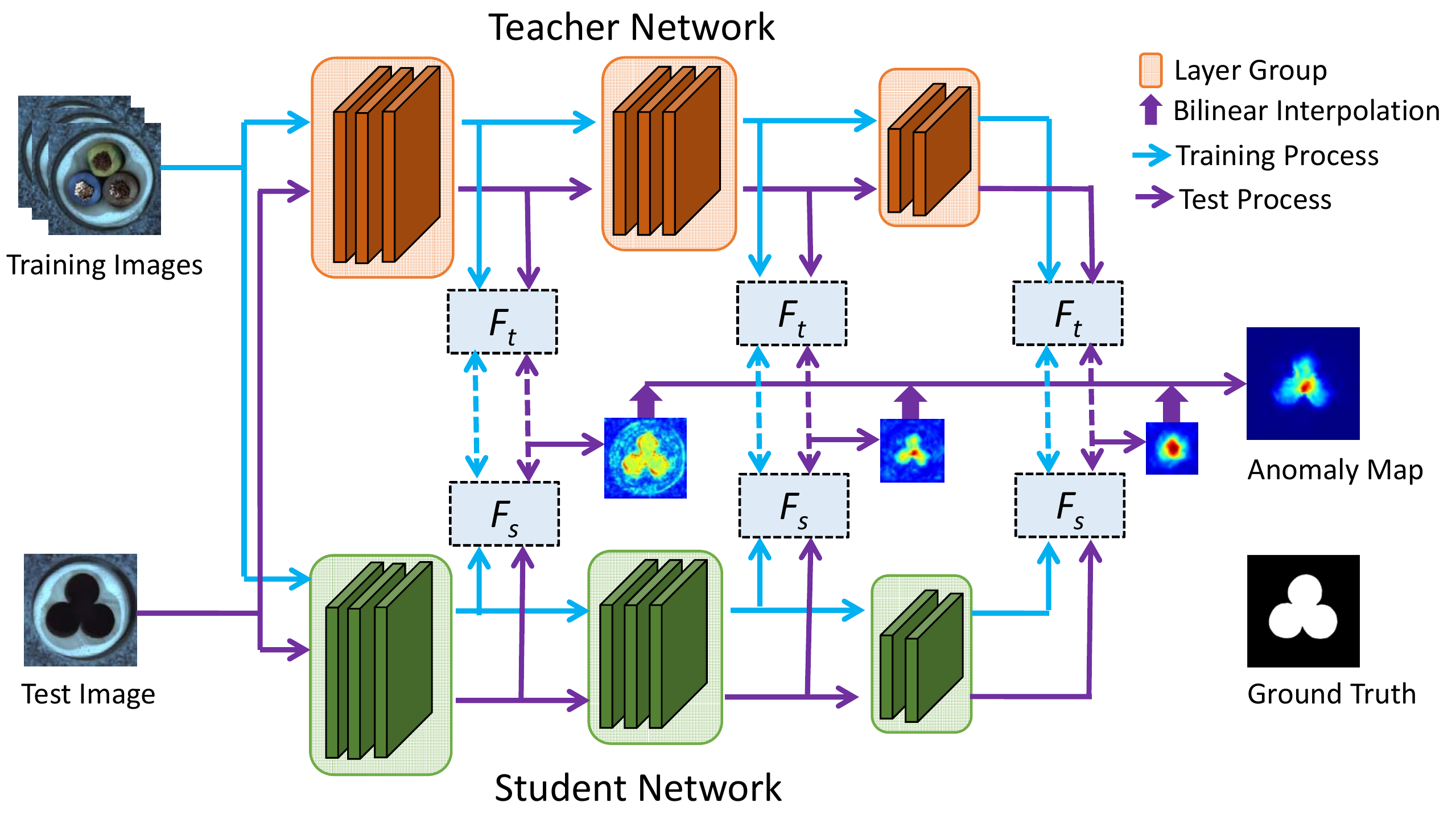}
	\caption{Schematic overview of our method. The feature pyramid of a student network is trained to match with the counterpart of a pre-trained teacher network. A test image (or pixel) has a high anomaly score if its features from the two models differ significantly. The feature pyramid matching enables our method to detect anomalies of various sizes with a single forward pass.}
	\label{fig2}
\end{figure*}

We make use of the student-teacher learning framework to implicitly model the feature distribution of the normal training images. The teacher is a powerful network pre-trained on the image classification task (\textit{e.g.}, a ResNet-18 pre-trained on ImageNet). To reduce information loss, the student shares the same architecture with the teacher. This is in essence one case of feature-based knowledge distillation~\cite{Wang2020}.

Here, we need to consider a key factor, \ie, position of distillation. Deep neural networks generate a pyramid of features for each input image. Bottom layers result in higher-resolution features encoding low-level information such as textures, edges and colors. By contrast, top layers yield low-resolution features that contain context information. The features created by bottom layers are often generic enough and they can be shared by various vision tasks~\cite{Oquab2014,Zeiler2014}. \wgd{This motivates us to integrate low-level and high-level features in a complementary way.} As different layers in deep neural networks correspond to distinct receptive fields, we select the features extracted by a few successive bottom layer groups (\textit{e.g.}, blocks in ResNet-18) of the teacher to guide the student's learning. This hierarchical feature matching allows our method to detect anomalies of various sizes.

Figure~\ref{fig2} gives a sketch of our method with the images from the MVTec AD dataset~\cite{Bergmann2020} as examples. The training and test processes are formally provided as follows.

\subsection{Training Process}
\label{sec3-sub1}

\wgd{The training phase aims to obtain a good student which can perfectly imitate the outputs of a fixed teacher on normal images. Formally, given} a training dataset of anomaly-free images $\mathcal{D} = \{\img{I}_1, \img{I}_2, \dots, \img{I}_n\}$, our goal is to capture the normal data manifold by \wgd{matching} the features extracted by the $L$ bottom layer groups of the teacher with the counterparts of the student. For an input image $\img{I}_k \in \R^{w \times h \times c}$, where $h$ is the height, $w$ is the width and $c$ is the number of the color channels, the $l$th bottom layer group of the teacher and student outputs a feature map $F^l_t(\img{I}_k) \in \R^{w_l \times h_l \times d_l}$ and $F^l_s(\img{I}_k) \in \R^{w_l \times h_l \times d_l}$, where $w_l$, $h_l$ and $d_l$ denote the width, height and channel number of the feature map, respectively. Since there is no prior knowledge regarding the appearances and locations of objects, we simply assume that all image regions are anomaly-free in the training set. Note that $F^l_t (\img{I}_k)_{ij} \in \R^{d_l}$ and  $F^l_s (\img{I}_k)_{ij} \in \R^{d_l}$ are feature vectors at position $(i, j)$ in the feature maps from the teacher and student, respectively. We define the loss at position $(i, j)$ as $\ell_2$-distance between the $\ell_2$-normalized feature vectors, namely,
\begin{equation}
\ell^{l}(\img{I}_k)_{ij} = \frac{1}{2} \left \|\hat{F}^l_t (\img{I}_k)_{ij} - \hat{F}^l_s (\img{I}_k)_{ij}\right\|^2_{\ell_2},
\label{eq1}
\end{equation}
\begin{equation}
\hat{F}^l_t (\img{I}_k)_{ij} = \frac{F^l_t (\img{I}_k)_{ij}}{\|F^l_t (\img{I}_k)_{ij}\|_{\ell_2}}, \,\,\, \hat{F}^l_s (\img{I}_k)_{ij} = \frac{F^l_s (\img{I}_k)_{ij}}{\|F^l_s (\img{I}_k)_{ij}\|_{\ell_2}}. \nonumber
\end{equation}

It is worth noting that the $\ell_2$ distance used in (Eq. \ref{eq1}) is \wgd{proportional} to the cosine \wgd{distance as $F^l_t(\img{I}_k)$ and $F^l_s(\img{I}_k)$ are $\ell_2$-normalized vectors}. Thus the loss $\ell^{l}(\img{I}_k)_{ij} \in (0, 1)$. The loss for the entire image $\img{I}_k$ is given as an average of the loss at each position,
\begin{equation}
\ell^{l}(\img{I}_k) = \frac{1}{w_lh_l} \sum_{i=1}^{w_l} \sum_{j=1}^{h_l} \ell^{l}(\img{I}_k)_{ij},
\label{eq2}
\end{equation}
and the total loss is the weighted average of the loss at different pyramid scales,
\begin{equation}
\ell (\img{I}_k) = \sum_{l=1}^L \alpha_l \ell^l (\img{I}_k), \quad \text{s.t.} \,\,\, \alpha_l \geq 0, \,\,\ 
\label{eq3}
\end{equation}
where $\alpha_l$ depicts the impact of the $l$th feature scale on anomaly detection. We simply set $\alpha_l = 1, l=1, \dots, L$ in all our experiments. Given a minibatch $\mathcal{B}$ sampled from the training dataset $\mathcal{D}$, we update the student by minimizing the loss $\ell_{\mathcal{B}} = \frac{1}{|\mathcal{B}|} \sum_{k \in \mathcal{B}} \ell (\img{I}_k)$. \wgd{Note that we only update the student while keeping the teacher fixed throughout the training phase.}

\subsection{Test Process}
\label{sec3-sub2}

In the test phase, we aim to obtain an anomaly map $\Omega$ of size $w \times h$ regarding a test image $\img{J} \in \R^{w \times h \times c}$. The score $\Omega_{ij} \in [0, 1]$ indicates how much the pixel at position $(i, j)$ deviates from the training data manifold. We forward the test image $\img{J}$ into the teacher and the student. Let $F^l_t(\img{J})$ and $F^l_s(\img{J})$ denote the feature maps generated by the $l$th bottom layer group of the teacher and the student, respectively. We can compute an anomaly map $\Omega^l (\img{J})$ of size $w_l \times h_l$, whose element $\Omega^l_{ij} (\img{J})$ is the loss (Eq. \ref{eq1}) at position $(i, j)$. The anomaly map $\Omega^l (\img{J})$ is upsampled to size $w \times h$ by bilinear interpolation. The resulting anomaly map is defined as the element-wise \wgd{product} of $L$ equal-sized upsampled anomaly maps,
\begin{equation}
\Omega(\img{J}) = \prod_{l=1}^L \upsample \Omega^l(\img{J}).
\label{eq4}
\end{equation}
A test image is designated as anomaly if any pixel in the image is anomalous. As a result, we simply choose the maximum value in the anomaly map, \ie,~$\max(\Omega (\img{J}))$ as the anomaly score for the test image $\img{J}$.

\section{Experiments}
\label{sec4}

\subsection{Dataset}
We conduct experiments on the MVTec Anomaly Detection (MVTec AD)~\cite{Bergmann2019} dataset, with both the image-level and pixel-level anomaly detection tasks considered. The dataset is specifically created to benchmark algorithms for anomaly localization. It collects more than 5,000 high-resolution images of industrial products covering 15 different categories. For each category, the training set only includes defect-free images and the test set comprises both defect-free images and defective images of different types. The performance is measured by two popular metrics: AUC-ROC and Per-Region-Overlap (PRO)~\cite{Bergmann2020}. Supplementary material provides more results on ShanghaiTech Campus (STC)~\cite{Luo_2017_ICCV} and CIFAR-10~\cite{Krizhevsky2009}.

\subsection{Implementation Details}

For all the experiments, we choose the first three blocks (\textit{i.e.}, conv2\_x, conv3\_x, conv4\_x) of ResNet-18 as the pyramid feature extractors for both the teacher and student networks. The parameters of the teacher network are copied from the ResNet-18 pre-trained on ImageNet, while those of the student network are initialized randomly. We train the network using stochastic gradient descent (SGD) with a learning rate of 0.4 for 100 epochs. The batch size is 32. All the images in the training and test sets are resized to 256$\times$256. 
\wgd{For each category, we use 80\% of training images to build the student, keeping the remaining 20\% for validation. We select the checkpoint with the lowest validation error (Eq.~\ref{eq1}) to perform anomaly detection.}

\subsection{Results}
We begin with the task of finding anomalous images. As defective regions usually occupy a small proportion of the whole image, the test anomalies differ in a subtle way from the training images. This makes the MVTec AD dataset more challenging than those previously used in the literature (\textit{e.g.}, MNIST and CIFAR-10) where the images from the other categories are regarded as anomalous to the selected one. Table~\ref{tab1} compares our method to state-of-the-art approaches: Geom~\cite{Golan2018}, GANomaly~\cite{Akcay2018}, $\l_2$-AE~\cite{Aytekin2018}, ITAE~\cite{Huang2020}, \wgd{Cut-Paste \cite{li2021cutpaste}} \wgd{Patch-SVDD \cite{yi2020patch}, PaDiM \cite{defard2021padim}} and SPADE~\cite{Cohen2020}. We clearly see that our approach outperforms all the other methods. In particular, the performance is improved up to 11.7\% compared with SPADE~\cite{Cohen2020}, which also leverages multi-scale features from a pre-trained model. It validates the superiority of the student-teacher learning framework. 

\begin{table*}[!t]
	\centering
	\begin{adjustbox}{width=0.95\textwidth}
	\begin{threeparttable}
		\begin{tabular}{c|c|c c c c c c c c c}
			\hline
			& Category   & SSIM-AE & AnoGAN & CNN-Dict$^*$ & STAD$^*$ & \wgd{Cut-Paste}    & \wgd{Patch-SVDD} & \wgd{PaDiM-R18$^*$} & SPADE$^*$  &  Ours$^*$\\  \hline
			\multirow{10}{*}{\rotatebox{90}{Textures}} 
			&\multirow{2}{*}{Carpet}  & 0.65  & 0.20  & 0.47  & 0.695 & \wgd{-}     & \wgd{-}     & \wgd{\textbf{0.960}} & 0.947 & 0.958 \\ 
			                     &    & 0.87  & 0.54  & 0.72  & -     & \wgd{0.983} & \wgd{0.926} & \wgd{\textbf{0.989}} & 0.975 & 0.988 \\ \cline{2-11} 
			&\multirow{2}{*}{Grid}    & 0.85  & 0.23  & 0.18  & 0.819 & \wgd{-}     & \wgd{-}     & \wgd{0.909} & 0.867 & \textbf{0.966} \\
			                     &    & 0.94  & 0.58  & 0.59  & -     & \wgd{0.975} & \wgd{0.962} & \wgd{0.949} & 0.937 & \textbf{0.990} \\ \cline{2-11} 
			&\multirow{2}{*}{Leather} & 0.56  & 0.38  & 0.64  & 0.819 & \wgd{-}     & \wgd{-}     & \wgd{0.979} & 0.972 & \textbf{0.980} \\
			                     &    & 0.78  & 0.64  & 0.87  & -     & \wgd{\textbf{0.995}} & \wgd{0.974} & \wgd{0.991} & 0.976 & 0.993 \\ \cline{2-11} 
			&\multirow{2}{*}{Tile}    & 0.18  & 0.18  & 0.80  & 0.912 & \wgd{-}     & \wgd{-}     & \wgd{0.816} & 0.759 & \textbf{0.921} \\
			                     &    & 0.59  & 0.50  & 0.93  & -     & \wgd{0.905} & \wgd{0.914} & \wgd{0.912} & 0.874 & \textbf{0.974} \\ \cline{2-11} 
			&\multirow{2}{*}{Wood}    & 0.61  & 0.39  & 0.62  & 0.725 & \wgd{-}     & \wgd{-}     & \wgd{0.903} & 0.874 & \textbf{0.936} \\ 
			                     &    & 0.73  & 0.62  & 0.91  & -     & \wgd{0.955} & \wgd{0.908} & \wgd{0.936} & 0.885 & \textbf{0.972} \\ \hline
			\multirow{20}{*}{\rotatebox{90}{Objects}}
			&\multirow{2}{*}{Bottle}  & 0.83  & 0.62  & 0.74  & 0.918 & \wgd{-}     & \wgd{ -}    & \wgd{0.939} & \textbf{0.955} & 0.951 \\
			                     &    & 0.93  & 0.86  & 0.78  & -     & \wgd{0.976} & \wgd{0.981} & \wgd{0.981} & 0.984 & \textbf{0.988} \\ \cline{2-11} 
			&\multirow{2}{*}{Cable}   & 0.48  & 0.38  & 0.56  & 0.865 & \wgd{-}     & \wgd{-}     & \wgd{0.862} & \textbf{0.909} & 0.877 \\
		                         &    & 0.82  & 0.78  & 0.79  & -     & \wgd{0.900} & \wgd{0.968} & \wgd{0.958} & \textbf{0.972} & 0.955 \\ \cline{2-11} 
			&\multirow{2}{*}{Capsule} & 0.86  & 0.31  & 0.31  & 0.916 & \wgd{-}     & \wgd{-}     & \wgd{0.919} & \textbf{0.937} & 0.922 \\
			                     &    & 0.94  & 0.84  & 0.84  & -     & \wgd{0.974} & \wgd{0.958} & \wgd{0.983} & \textbf{0.990} & 0.983 \\ \cline{2-11} 
			&\multirow{2}{*}{Hazelnut}& 0.92  & 0.70  & 0.84  & 0.937 & \wgd{-}     & \wgd{-}     & \wgd{0.914} & \textbf{0.954} & 0.943 \\
			                     &    & 0.97  & 0.87  & 0.72  & -     & \wgd{0.973} & \wgd{0.975} & \wgd{0.977} & \textbf{0.991} & 0.985 \\ \cline{2-11} 
			&\multirow{2}{*}{Metal nut}&0.60  & 0.32  & 0.36  & 0.895 & \wgd{-}     & \wgd{-}     & \wgd{0.819} & 0.944 & \textbf{0.945} \\
			                     &    & 0.89  & 0.76  & 0.82  & -     & \wgd{0.931} & \wgd{0.980} & \wgd{0.967} & \textbf{0.981} & 0.976 \\ \cline{2-11} 
			&\multirow{2}{*}{Pill}    & 0.83  & 0.78  & 0.46  & 0.935 & \wgd{-}     & \wgd{-}     & \wgd{0.906} & 0.946 & \textbf{0.965} \\
			                     &    & 0.91  & 0.87  & 0.68  & -     & \wgd{0.957} & \wgd{0.951} & \wgd{0.947} & 0.965 & \textbf{0.978} \\ \cline{2-11} 
			&\multirow{2}{*}{Screw}   & 0.89  & 0.47  & 0.28  & 0.928 & \wgd{-}     & \wgd{-}     & \wgd{0.913} & \textbf{0.960} & 0.930 \\
			                     &    & 0.96  & 0.80  & 0.87  & -     & \wgd{0.967} & \wgd{0.957} & \wgd{0.974} & \textbf{0.989} & 0.983 \\ \cline{2-11}
			&\multirow{2}{*}{Toothbrush}& 0.78 & 0.75 & 0.15  & 0.863 & \wgd{-}     & \wgd{-}     & \wgd{0.923} & \textbf{0.935} & 0.922 \\   
			                     &    & 0.92  & 0.93  & 0.90  & -     & \wgd{0.981} & \wgd{0.981} & \wgd{0.987} & 0.979 & \textbf{0.989} \\ \cline{2-11}
			&\multirow{2}{*}{Transistor}& 0.73 & 0.55 & 0.63  & 0.701 & \wgd{-}     & \wgd{-}     & \wgd{0.802} & \textbf{0.874} & 0.695 \\
			                     &    & 0.90  & 0.86  & 0.66  & -     & \wgd{0.930} & \wgd{0.970} & \wgd{\textbf{0.972}} & 0.941 & 0.825 \\ \cline{2-11}
			&\multirow{2}{*}{Zipper}  & 0.67  & 0.47  & 0.70  & 0.933 & \wgd{-}     & \wgd{-}     & \wgd{0.947} & 0.926   &\textbf{0.952} \\
			                     &    & 0.88  & 0.78  & 0.76  & -     & \wgd{\textbf{0.993}} & \wgd{0.951} & \wgd{0.982} & 0.965   &  0.985 \\ \cline{2-11}
			&\multirow{2}{*}{Mean}    & 0.69  & 0.44  & 0.52  & 0.857 & \wgd{-}     & \wgd{-}     & \wgd{0.901} & 0.917   & \textbf{0.921} \\
			                     &    & 0.87  & 0.74  & 0.78  & -     & \wgd{0.960} & \wgd{0.957} & \wgd{0.967} & 0.965   & \textbf{0.970} \\
			\hline
		\end{tabular}
	\begin{tablenotes}
		\footnotesize
		  \item  $^*$ denotes extra dataset pre-trained model used.
		\end{tablenotes}
	\end{threeparttable}
	\end{adjustbox}
	
	\caption{Pixel-level anomaly detection. For each dataset category, PRO (top row) and AUC-ROC (bottom row) scores are given.}
	\label{tab:summary}
\end{table*}

\begin{table}[!t]
	\centering
	\begin{threeparttable}
	\setlength{\tabcolsep}{1.5mm}{
	\begin{adjustbox}{width=0.98\textwidth}
		\begin{tabular}{ccccccccc}
			\hline
			Geom  & GANomaly & $\l_2$-AE & ITAE & \wgd{Cut-Paste} & \wgd{Patch-SVDD} & \wgd{PaDiM-WR50$^*$} & SPADE$^*$ & Ours\\
			\hline
			0.672 & 0.762  & 0.754  & 0.839 & \wgd{0.952} & \wgd{0.921} & \wgd{0.953} & 0.855 & \textbf{0.955} \\
			\hline
		\end{tabular}
		\end{adjustbox}
		}
		\begin{tablenotes}
		\footnotesize
		  \item  $^*$ denotes extra dataset pre-trained model used.
		\end{tablenotes}
	\end{threeparttable}
	\caption{Image-level anomaly detection. The performance is measured by average AUC-ROC across 15 categories.}
	\label{tab1}
\end{table}

We then consider the task of pixel-level anomaly detection and compare our method with the counterparts \wgd{including Patch-SVDD \cite{yi2020patch}, PaMiD \cite{defard2021padim}, \emph{etc}}. Table~\ref{tab:summary} reports the performance in terms of the AUC-ROC and PRO metrics. We notice two trends to achieve performance gains: (1) by pre-trained models, with a Wide-ResNet50$\times$2 network \cite{zagoruyko2016wide}, SPADE reports very competitive scores; (2) by self-training techniques, Cut-Paste \cite{li2021cutpaste} and Patch-SVDD \cite{yi2020patch} show this potential through designing proper pretext tasks for feature learning. As our approach assumes that anomaly detection is fulfilled via the heterogeneity of the student and teacher networks, \emph{i.e.} different network parameters learned from individual data, we employ a pre-trained model built on generic images rather than self-supervised learning on the small scale anomaly detection dataset. As Table~\ref{tab:summary} displays, our approach delivers better performance than the others. It should be noted although STAD~\cite{Bergmann2020} adopts the student-teacher learning framework, its performance is always inferior to that of our method. This gap can be attributed to the information loss in its two-step and single-scale knowledge transfer process. This validates our improvement in feature learning. When equipped with the same backbone as SPADE~\cite{Cohen2020}, our method further boosts the results, \emph{i.e.} 0.973 and 0.923 in AUC-ROC and PRO, respectively.


\section{Ablation Studies and Discussions}
\label{sec5}

\wgd{We first perform feature visualization to investigate what the student learns from its teacher and also conduct} ablation studies on the MVTec AD dataset to answer the following three questions. Is feature pyramid matching superior to single feature \wgd{matching}? Is the teacher pre-trained on other datasets still useful? \wgd{Is our method applicable to small training dataset?}

\subsection{Feature Visualization}

\begin{figure}
	\centering
	\subfigure[Bottle (broken$\_$large)]{
	    \begin{minipage}[b]{0.28\textwidth}
	        \includegraphics[width=3.6cm,height=3.6cm]{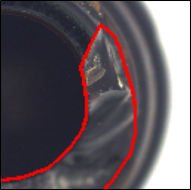}
	    \end{minipage}
	}
	\subfigure[Points at various positions]{
	    \begin{minipage}[b]{0.28\textwidth}
	        \includegraphics[width=3.6cm,height=3.6cm]{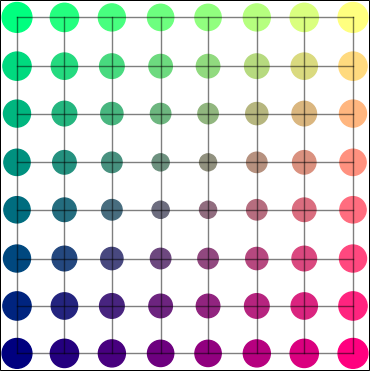}
	    \end{minipage}
	}
	\subfigure[t-SNE]{
	    \begin{minipage}[b]{0.35\textwidth}
	        \includegraphics[width=4.5cm,height=3.6cm]{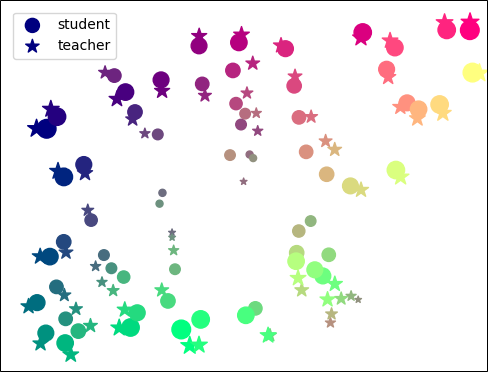}
	    \end{minipage}
	}
	\caption{$t$-SNE visualization \cite{van2008visualizing} of learned features from the student and teacher. (a) an example test image with defects contoured by a red line. (b) point map in which different positions are encoded by different sizes and colors. (c) $t$-SNE visualizations for features from the student (circle) and the teacher (star) with (a) as input. Zoomed in for better display.}
	\label{fig:feature}
\end{figure}

\wgd{Figure \ref{fig:feature} shows $t$-SNE visualization \cite{van2008visualizing} of learned features from the student and teacher. 
Obviously, the features from the student and teacher on normal regions distribute closer (even overlapped) than the ones on anomalous regions. It suggests that the student learns to match the teacher's output on normal images. It also shows that the student well captures the distribution of normal patterns under the supervision of a good teacher.}

\subsection{Feature Matching}
\label{sec5-sub1}

We first minutely investigate the effectiveness of feature extraction by each individual block of ResNet-18. Considering that the first block is a simple convolutional layer, we exclude it from comparison. We train the student by matching features extracted by its second, third, fourth and fifth blocks with the counterparts of the teacher respectively. As shown in Table~\ref{tab6}, feature \wgd{matching} conducted at the end of the third and fourth blocks can achieve better performance. This is in good agreement with the previous discovery that the middle-level features play a more important role in knowledge transfer~\cite{Oquab2014}.

We then test three different combinations of the consecutive blocks of ResNet-18. Likewise, we \wgd{match} the features extracted from the corresponding compound blocks of the teacher and the student. Table~\ref{tab6} shows that the mixture of the second, third and fourth blocks outperforms other combinations as well as the single components. It implies that feature pyramid matching is a better way for feature learning. This finding is also validated in Figure~\ref{fig1}. Anomaly maps generated by low-level features are more suitable for precise anomaly localization, but they are likely to include background noise. By contrast, anomaly maps generated by high-level features are able to segment big anomalous regions. The aggregation of anomaly maps at different scales contributes to accurate detection of anomalies of various sizes.

\begin{table}[!]
	\centering
	\begin{adjustbox}{width=0.9\textwidth}
		\begin{tabular}{c|ccccccc}
			\hline
			\diagbox{Metric}{\# Block} & 2 & 3 & 4 & 5 & [2, 3] & [2, 3, 4] & [2, 3, 4, 5]\\
			\hline
			AR$_I$ & 0.808 & 0.917 & 0.934 & 0.819 & 0.849 & \textbf{0.955} & 0.949\\
			AR$_P$ & 0.915 & 0.953 & 0.957 & 0.860 & 0.950 & \textbf{0.970} &  0.969\\
			PRO & 0.815 & 0.897 & 0.835 & 0.504 & 0.886 & \textbf{0.921} & 0.886\\
			\hline
		\end{tabular}
	\end{adjustbox}
	\caption{Ablation studies for feature matching. The performance is measured by the average image-level AUC-ROC (AR$_I$), average pixel-level AUC-ROC (AR$_P$) and average PRO across 15 categories.}
	\label{tab6}
\end{table}

\subsection{Pre-trained Datasets}
\label{sec5-sub3}

\begin{table}[!]
	\centering
	\begin{adjustbox}{width=0.8\textwidth}
		\begin{tabular}{c|cccccc}
			\hline
			\diagbox{Metric}{Dataset} & ImageNet & MNIST & CIFAR-10 & CIFAR-100 & SVHN\\
			\hline
			AR$_I$ & \textbf{0.955} & 0.619 & 0.826 & 0.835 & 0.796\\
			AR$_P$ & \textbf{0.970} & 0.759 & 0.931 & 0.937 & 0.902\\
			PRO & \textbf{0.921} & 0.528  & 0.863 & 0.842 & 0.742\\
			\hline
		\end{tabular}
	\end{adjustbox}
	\caption{Ablation studies for pre-trained datasets. The performance is measured by the average image-level AUC-ROC (AR$_I$), average pixel-level AUC-ROC (AR$_P$) and average PRO across 15 categories.}
	\label{tab8}
\end{table}

To answer the second question, we pre-train the teacher on a couple of image classification benchmarks, including MNIST \cite{yann2010mnist}, CIFAR-10 \cite{Krizhevsky2009}, CIFAR-100~\cite{Krizhevsky2009}, and SVHN~\cite{Netzer2011}. These pre-trained teachers are individually exploited to guide the student training. The MNIST and SVHN datasets simply contain digital numbers from 0 to 9. We see from Table~\ref{tab8} that the teacher networks pre-trained on these two datasets yield worse results. It indicates that the features learned from these two pre-trained models generalize poorly on the MVTec AD dataset. By contrast, the features extracted from the teacher networks pre-trained on CIFAR-10 and CIFAR-100 exhibit better generalization, as they contain more natural images. Note that the performance of these two pre-trained teachers is still inferior to that of the teacher pre-trained on ImageNet. This is because that the ImageNet dataset consists of a huge number of high-resolution natural images, which is crucial to learning more discriminating features.

\subsection{Number of Training Samples}
\label{sec5:sub4}
We investigate the effect of the training set size in this experiment. Only 5\% and 10\% anomaly-free images are used to train our model. It can be seen in Table~\ref{tab9} that our model still reaches a satisfactory level even if only a few training images are available. By contrast, SPADE suffers a serious performance degradation. This is caused by the missing of the tailored feature learning. Our model profits from this strategy and can capture the feature distribution of anomaly-free images in the few-shot scenario. Furthermore, our method uses only 10\% training samples to outperform the preliminary student-teacher framework \cite{Bergmann2020}. It validates the effectiveness of our feature pyramid matching technique.

\begin{table}[!]
	\centering
	\begin{adjustbox}{width=0.5\textwidth}
		\begin{tabular}{c|cc|cc}
			\hline
			& \multicolumn{2}{c|}{5\%} & \multicolumn{2}{c}{10\%} \\
			\hline
			Metric & Ours & SPADE & Ours & SPADE \\
			\hline
			AR$_I$ & \textbf{0.871} & 0.782 & \textbf{0.907} & 0.797 \\
			AR$_P$ & \textbf{0.961} & 0.932 & \textbf{0.967} & 0.955 \\
			PRO & \textbf{0.892} & 0.842 & \textbf{0.913} & 0.890 \\
			\hline
		\end{tabular}
	\end{adjustbox}
	\caption{Performance in terms of the number of training samples. The performance is measured by the average image-level AUC-ROC (AR$_I$), average pixel-level AUC-ROC (AR$_P$) and average PRO across 15 categories.}
	\label{tab9}
\end{table}

\section{Conclusion}
\label{sec6}

We present a new feature pyramid matching technique and incorporate it into the student-teacher anomaly detection framework. Given a powerful network pre-trained on image classification as the teacher, we use its different levels of features to guide a student network with the same structure to learn the distribution of anomaly-free images. On account of the hierarchical feature \wgd{matching}, our method is capable of detecting anomalies of various sizes with only a single forward pass. 
Experimental results on the MVTec AD dataset show that our method achieves superior performance to the state-of-the-art.

\section*{Acknowledgment}
This work is supported by the National Natural Science Foundation of China (62022011), the Research Program of State Key Laboratory of Software Development Environment (SKLSDE-2021ZX-04), and the Fundamental Research Funds for the Central Universities.

\bibliography{egbib}

\begin{thebibliography}{48}
\providecommand{\natexlab}[1]{#1}
\providecommand{\url}[1]{\texttt{#1}}
\expandafter\ifx\csname urlstyle\endcsname\relax
  \providecommand{\doi}[1]{doi: #1}\else
  \providecommand{\doi}{doi: \begingroup \urlstyle{rm}\Url}\fi

\bibitem[Abati et~al.(2019)Abati, Porrello, Calderara, and
  Cucchiara]{Abati2019}
Davide Abati, Angelo Porrello, Simone Calderara, and Rita Cucchiara.
\newblock Latent space autoregression for novelty detection.
\newblock In \emph{CVPR}, 2019.

\bibitem[Akcay et~al.(2018)Akcay, Atapour-Abarghouei, and Breckon]{Akcay2018}
Samet Akcay, Amir Atapour-Abarghouei, and Toby~P. Breckon.
\newblock {GAN}omaly: Semi-supervised anomaly detection via adversarial
  training.
\newblock In \emph{ACCV}, 2018.

\bibitem[An and Cho(2015)]{An2015}
Jinwon An and Sungzoon Cho.
\newblock Variational autoencoder based anomaly detection using reconstruction
  probabiliy.
\newblock Technical report, SNU Data Mining Center, 2015.

\bibitem[Andrews et~al.(2016)Andrews, Tanay, Morton, and Griffin]{Andrews2016}
Jerone T.~A. Andrews, Thomas Tanay, Edward~J. Morton, and Lewis~D. Griffin.
\newblock Transfer representation-learning for anomaly detection.
\newblock In \emph{ICML Workshops}, 2016.

\bibitem[Aytekin et~al.(2018)Aytekin, Ni, Cricri, and Aksu]{Aytekin2018}
Caglar Aytekin, Xingyang Ni, Francesco Cricri, and Emre Aksu.
\newblock Clustering and unsupervised anomaly detection with $l_2$ normalized
  deep auto-encoder representations.
\newblock In \emph{IJCNN}, 2018.

\bibitem[Baur et~al.(2018)Baur, Wiestler, Albarqouni, and Navab]{Baur2018}
Christoph Baur, Benedikt Wiestler, Shadi Albarqouni, and Nassir Navab.
\newblock Deep autoencoding models for unsupervised anomaly segmentation in
  brain mr images.
\newblock In \emph{MICCAI Workshops}, 2018.

\bibitem[Bergmann et~al.(2019)Bergmann, Fauser, Sattlegger, and
  Steger]{Bergmann2019}
Paul Bergmann, Michael Fauser, David Sattlegger, and Carsten Steger.
\newblock Mvtec {AD} - {A} comprehensive real-world dataset for unsupervised
  anomaly detection.
\newblock In \emph{CVPR}, 2019.

\bibitem[Bergmann et~al.(2020)Bergmann, Fauser, Sattlegger, and
  Steger]{Bergmann2020}
Paul Bergmann, Michael Fauser, David Sattlegger, and Carsten Steger.
\newblock Uninformed students: Student-teacher anomaly detection with
  discriminative latent embeddings.
\newblock In \emph{CVPR}, 2020.

\bibitem[Burlina et~al.(2019)Burlina, Joshi, and Wang]{Burlina2019}
Philippe Burlina, Neil Joshi, and I-Jeng Wang.
\newblock Where's wally now? deep generative and discriminative embeddings for
  novelty detection.
\newblock In \emph{CVPR}, 2019.

\bibitem[Chalapathy et~al.(2018)Chalapathy, Menon, and Chawla]{Chalapathy2018}
Raghavendra Chalapathy, Aditya~Krishna Menon, and Sanjay Chawla.
\newblock Anomaly detection using one-class neural networks.
\newblock \emph{arXiv:1802.06360}, 2018.

\bibitem[Cohen and Hoshen(2020)]{Cohen2020}
Niv Cohen and Yedid Hoshen.
\newblock Sub-image anomaly detection with deep pyramid correspondences.
\newblock \emph{arXiv:2005.02357}, 2020.

\bibitem[Deecke et~al.(2018)Deecke, Vandermeulen, Mandt, and Kloft]{Deecke2018}
Lucas Deecke, Robert Vandermeulen, Lukas~RuffStephan Mandt, and Marius Kloft.
\newblock Image anomaly detection with generative adversarial networks.
\newblock In \emph{ECML-PKDD}, pages 3--17, 2018.

\bibitem[Defard et~al.(2021)Defard, Setkov, Loesch, and
  Audigier]{defard2021padim}
Thomas Defard, Aleksandr Setkov, Angelique Loesch, and Romaric Audigier.
\newblock Padim: A patch distribution modeling framework for anomaly detection
  and localization.
\newblock In \emph{ICPR}, 2021.

\bibitem[Erfani et~al.(2016)Erfani, Rajasegarar, Karunasekera, and
  Leckie]{Erfani2016}
Sarah~M. Erfani, Sutharshan Rajasegarar, Shanika Karunasekera, and Christopher
  Leckie.
\newblock High-dimensional and large-scale anomaly detection using a linear
  one-class svm with deep learning.
\newblock \emph{Pattern Recognit.}, 58:\penalty0 121--134, 2016.

\bibitem[Golan and El{-}Yaniv(2018)]{Golan2018}
Izhak Golan and Ran El{-}Yaniv.
\newblock Deep anomaly detection using geometric transformations.
\newblock In \emph{NeurIPS}, 2018.

\bibitem[He et~al.(2016)He, Zhang, Ren, and Sun]{He2016}
Kaiming He, Xiangyu Zhang, Shaoqing Ren, and Jian Sun.
\newblock Deep residual learning for image recognition.
\newblock In \emph{CVPR}, 2016.

\bibitem[Huang et~al.(2020)Huang, Ye, Cao, Li, Zhang, and Lu]{Huang2020}
Chaoqin Huang, Fei Ye, Jinkun Cao, Maosen Li, Ya~Zhang, and Cewu Lu.
\newblock Attribute restoration framework for anomaly detection.
\newblock \emph{arXiv:1911.10676}, 2020.

\bibitem[Krizhevsky and Hinton(2009)]{Krizhevsky2009}
Alex Krizhevsky and Geoffrey Hinton.
\newblock Learning multiple layers of features from tiny images.
\newblock Technical report, University of Toronto, 2009.

\bibitem[LeCun and Cortes(2010)]{yann2010mnist}
Yann LeCun and Corinna Cortes.
\newblock Mnist handwritten digit database.
\newblock 2010.

\bibitem[LeCun et~al.(1998)LeCun, Bottou, Bengio, and Haffner]{Lecun1998}
Yann LeCun, L\'{e}on Bottou, Yoshua Bengio, and Patrick Haffner.
\newblock Gradient-based learning applied to document recognition.
\newblock \emph{Proc. IEEE}, 86\penalty0 (11):\penalty0 2278--2324, 1998.

\bibitem[{Li} et~al.(2021){Li}, {Sohn}, {Yoon}, and {Pfister}]{li2021cutpaste}
Chun-Liang {Li}, Kihyuk {Sohn}, Jinsung {Yoon}, and Tomas {Pfister}.
\newblock Cutpaste: Self-supervised learning for anomaly detection and
  localization.
\newblock In \emph{CVPR}, 2021.

\bibitem[Liu et~al.(2020)Liu, Li, Zheng, Karanam, Wu, Bhanu, Radke, and
  Camps]{liu2020towards}
Wenqian Liu, Runze Li, Meng Zheng, Srikrishna Karanam, Ziyan Wu, Bir Bhanu,
  Richard~J Radke, and Octavia Camps.
\newblock Towards visually explaining variational autoencoders.
\newblock In \emph{CVPR}, 2020.

\bibitem[Luo et~al.(2017)Luo, Liu, and Gao]{Luo_2017_ICCV}
Weixin Luo, Wen Liu, and Shenghua Gao.
\newblock A revisit of sparse coding based anomaly detection in stacked rnn
  framework.
\newblock In \emph{ICCV}, 2017.

\bibitem[Masana et~al.(2018)Masana, Ruiz, Serrat, Joost, and Lopez]{Masana2018}
Marc Masana, Idoia Ruiz, Joan Serrat, Van De~Weijer Joost, and Antonio~M Lopez.
\newblock Metric learning for novelty and anomaly detection.
\newblock In \emph{BMVC}, 2018.

\bibitem[Moya et~al.(1993)Moya, Koch, and Hostetler]{Moya1993}
M.~M. Moya, M.~W. Koch, and L.~D. Hostetler.
\newblock One-class classifier networks for target recognition applications.
\newblock In \emph{WCCI}, 1993.

\bibitem[Nalisnick et~al.(2019)Nalisnick, Matsukawa, Teh, Gorur, and
  Lakshminarayanan]{Nalisnick2019}
Eric Nalisnick, Akihiro Matsukawa, Yee~Whye Teh, Dilan Gorur, and Balaji
  Lakshminarayanan.
\newblock Do deep generative models know what they don't know?
\newblock In \emph{ICLR}, 2019.

\bibitem[Napoletano et~al.(2018)Napoletano, Piccoli, and
  Schettini]{Napoletano2018}
Paolo Napoletano, Flavio Piccoli, and Raimondo Schettini.
\newblock Anomaly detection in nanofibrous materials by {CNN}-based
  self-similarity.
\newblock \emph{Sensors}, 18\penalty0 (2):\penalty0 209, 2018.

\bibitem[Netzer et~al.(2011)Netzer, Wang, Coates, Bissacco, Wu, and
  Ng]{Netzer2011}
Yuval Netzer, Tao Wang, Adam Coates, Alessandro Bissacco, Bo~Wu, and Andrew~Y.
  Ng.
\newblock Reading digits in natural images with unsupervised feature learning.
\newblock In \emph{NeurIPS Workshops}, 2011.

\bibitem[Oquab et~al.(2014)Oquab, Bottou, Laptev, and Sivic]{Oquab2014}
Maxime Oquab, L\'{e}on Bottou, Ivan Laptev, and Josef Sivic.
\newblock Learning and transferring mid-level image representations using
  convolutional neural networks.
\newblock In \emph{CVPR}, 2014.

\bibitem[Paul~Bergmann and Steger(2019)]{Bergmann2019b}
Michael Fauser David~Sattlegger Paul~Bergmann, Sindy~L\"{o}we and Carsten
  Steger.
\newblock Improving unsupervised defect segmentation by applying structural
  similarity to autoencoders.
\newblock In \emph{VISIGRAPP}, 2019.

\bibitem[Roitberg et~al.(2018)Roitberg, Al-Halah, and
  Stiefelhagen]{Roitberg2018}
Alina Roitberg, Ziad Al-Halah, and Rainer Stiefelhagen.
\newblock Informed democracy: Voting-based novelty detection for action
  recognition.
\newblock In \emph{BMVC}, 2018.

\bibitem[Ruff et~al.(2018)Ruff, Vandermeulen, G\"{o}rnitz, Deecke, Siddiqui,
  Binder, M\"{u}ller, and Kloft]{Ruff2018}
Lukas Ruff, Robert~A. Vandermeulen, Nico G\"{o}rnitz, Lucas Deecke, Shoaib~A.
  Siddiqui, Alexander Binder, Emmanuel M\"{u}ller, and Marius Kloft.
\newblock Deep one-class classification.
\newblock In \emph{ICML}, 2018.

\bibitem[Sabokrou et~al.(2018)Sabokrou, Fayyaz, Fathy, Moayed, and
  Klette]{Sabokrou2018}
Mohammad Sabokrou, Mohsen Fayyaz, Mahmood Fathy, Zahra Moayed, and Reinhard
  Klette.
\newblock Deep-anomaly: Fully convolutional neural network for fast anomaly
  detection in crowded scenes.
\newblock \emph{CVIU}, 172, 2018.

\bibitem[Schlegl et~al.(2017)Schlegl, Seeb\"{o}ck, Waldstein, Schmidt-Erfurth,
  and Langs]{Schlegl2017}
Thomas Schlegl, Philipp Seeb\"{o}ck, Sebastian~M. Waldstein, Ursula
  Schmidt-Erfurth, and Georg Langs.
\newblock Unsupervised anomaly detection with generative adversarial networks
  to guide marker discovery.
\newblock In \emph{IPMI}, 2017.

\bibitem[Schlegl et~al.(2019)Schlegl, Seeb\"{o}ck, Waldstein, Langs, and
  Schmidt-Erfurth]{Schlegl2019}
Thomas Schlegl, Philipp Seeb\"{o}ck, Sebastian~M. Waldstein, Georg Langs, and
  Ursula Schmidt-Erfurth.
\newblock f-{AnoGAN}: Fast unsupervised anomaly detection with generative
  adversarial networks.
\newblock \emph{MED IMAGE ANAL}, 54:\penalty0 30--44, 2019.

\bibitem[Sch\"{o}lkopf et~al.(2001)Sch\"{o}lkopf, Platt, Shawe-Taylor, Smola,
  and Williamson]{Scholkopf2001}
Bernhard Sch\"{o}lkopf, John~C. Platt, John Shawe-Taylor, Alex~J. Smola, and
  Robert~C. Williamson.
\newblock Estimating the support of a high-dimensional distribution.
\newblock \emph{NEURAL COMPUT}, 13\penalty0 (7), 2001.

\bibitem[Seeb\"{o}ck et~al.(2016)Seeb\"{o}ck, Waldstein, Klimscha, Donner,
  Schlegl, Schmidt-Erfurth, and Langs]{Seebock2016}
Philipp Seeb\"{o}ck, Sebastian Waldstein, Sophie Klimscha, Bianca S.
  Gerendas~Ren\'{e} Donner, Thomas Schlegl, Ursula Schmidt-Erfurth, and Georg
  Langs.
\newblock Identifying and categorizing anomalies in retinal imaging data.
\newblock \emph{arXiv:1612.00686}, 2016.

\bibitem[Selvaraju et~al.(2017)Selvaraju, Cogswell, Das, Vedantam, and
  Parikh]{Selvaraju2017}
Ramprasaath~R. Selvaraju, Michael Cogswell, Abhishek Das, Ramakrishna Vedantam,
  and Devi Parikh.
\newblock {Grad-CAM}: Visual explanations from deep networks via gradient-based
  localization.
\newblock In \emph{ICCV}, 2017.

\bibitem[Van~der Maaten and Hinton(2008)]{van2008visualizing}
Laurens Van~der Maaten and Geoffrey Hinton.
\newblock Visualizing data using t-sne.
\newblock \emph{JMLR}, 9\penalty0 (11), 2008.

\bibitem[Vasilev et~al.(2018)Vasilev, Golkov, Lipp, Sgarlata, Tomassini, Jones,
  and Cremers]{Vasilev2018}
Aleksei Vasilev, Vladimir Golkov, Ilona Lipp, Eleonora Sgarlata, Valentina
  Tomassini, Derek~K. Jones, and Daniel Cremers.
\newblock {q-Space} novelty detection with variational autoencoders.
\newblock \emph{arXiv:1806.02997}, 2018.

\bibitem[Venkataramanan et~al.(2020)Venkataramanan, Peng, Singh, and
  Mahalanobis]{shashanka2020attention}
Shashanka Venkataramanan, Kuan-Chuan Peng, Rajat~Vikram Singh, and Abhijit
  Mahalanobis.
\newblock Attention guided anomaly localization in images.
\newblock In \emph{ECCV}, 2020.

\bibitem[Wang and Yoon(2020)]{Wang2020}
Lin Wang and Kuk-Jin Yoon.
\newblock Knowledge distillation and student-teacher learning for visual
  intelligence: A review and new outlooks.
\newblock \emph{arXiv preprint arXiv:2004.05937}, 2020.

\bibitem[Yi and Yoon(2020)]{yi2020patch}
Jihun Yi and Sungroh Yoon.
\newblock Patch svdd: Patch-level svdd for anomaly detection and segmentation.
\newblock In \emph{ACCV}, 2020.

\bibitem[Zagoruyko and Komodakis(2016)]{zagoruyko2016wide}
Sergey Zagoruyko and Nikos Komodakis.
\newblock Wide residual networks.
\newblock \emph{arXiv preprint arXiv:1605.07146}, 2016.

\bibitem[Zeiler and Fergus(2014)]{Zeiler2014}
Matthew~D. Zeiler and Rob Fergus.
\newblock Visualizing and understanding convolutional networks.
\newblock In \emph{ECCV}, 2014.

\bibitem[Zenati et~al.(2018)Zenati, Romain, Foo, Lecouat, and
  Chandrasekhar]{Zenati2018}
Houssam Zenati, Manon Romain, Chuan-Sheng Foo, Bruno Lecouat, and Vijay
  Chandrasekhar.
\newblock Adversarially learned anomaly detection.
\newblock In \emph{ICDM}, 2018.

\bibitem[Zhou and Paffenroth(2017)]{Zhou2017}
Chong Zhou and Randy~C. Paffenroth.
\newblock Anomaly detection with robust deep autoencoders.
\newblock In \emph{KDD}, 2017.

\bibitem[Zong et~al.(2018)Zong, Song, Min, Cheng, Lumezanu, Cho, and
  Chen]{Zong2018}
Bo~Zong, Qi~Song, Martin~Renqiang Min, Wei Cheng, Cristian Lumezanu, Daeki Cho,
  and Haifeng Chen.
\newblock Deep autoencoding gaussian mixture model for unsupervised anomaly
  detection.
\newblock In \emph{ICLR}, 2018.

\end{thebibliography}
\end{document}